\documentclass[conference]{IEEEtran}
\usepackage{cite}
\ifCLASSINFOpdf
\else
\fi
\usepackage[cmex10]{amsmath}
\usepackage{algorithmic}
\usepackage{amsmath,bm}
\usepackage{amssymb}
\usepackage{algorithm}

\usepackage{graphicx}
\graphicspath{{./fig/}}
\begin{document}
\title{Evidential relational clustering  using medoids}

\iffalse
\author{\IEEEauthorblockN{Michael Shell}
\IEEEauthorblockA{School of Electrical and\\Computer Engineering\\
Georgia Institute of Technology\\
Atlanta, Georgia 30332--0250\\
Email: http://www.michaelshell.org/contact.html}
\and
\IEEEauthorblockN{Homer Simpson}
\IEEEauthorblockA{Twentieth Century Fox\\
Springfield, USA\\
Email: homer@thesimpsons.com}
\and
\IEEEauthorblockN{James Kirk\\ and Montgomery Scott}
\IEEEauthorblockA{Starfleet Academy\\
San Francisco, California 96678--2391\\
Telephone: (800) 555--1212\\
Fax: (888) 555--1212}}
\fi



\author{\IEEEauthorblockN{Kuang Zhou$^\text{a,b}$,
Arnaud Martin$^\text{b}$,
Quan Pan$^\text{a}$, and
Zhun-ga Liu$^\text{a}$}
\IEEEauthorblockA{a. School of Automation, Northwestern Polytechnical University,
Xi'an, Shaanxi 710072, PR China. }
\IEEEauthorblockA{b. DRUID, IRISA, University of Rennes 1, Rue E. Branly, 22300 Lannion, France}
}


\maketitle

\begin{abstract}
In real clustering applications, proximity data,  in which only pairwise similarities or dissimilarities are known, is more general than object data,  in which each pattern is described explicitly by a list of attributes.
Medoid-based clustering algorithms, which assume the prototypes of classes are objects, are of great value for partitioning relational data sets. In this paper a new prototype-based clustering method, named Evidential $C$-Medoids (ECMdd), which is an extension of Fuzzy $C$-Medoids (FCMdd) on the theoretical framework of belief functions is proposed.  In ECMdd,  medoids are utilized as the  prototypes to represent the detected classes, including specific classes and imprecise classes. Specific classes are for the data  which are distinctly far from the prototypes of other classes, while  imprecise classes accept the objects that may be close to the prototypes of more than one class.  This soft decision mechanism could make the clustering results more cautious and reduce the misclassification rates.  Experiments in  synthetic and   real data sets are used to illustrate the performance of ECMdd. The results show  that ECMdd could capture well the uncertainty in the internal data structure. Moreover, it is more robust to the  initializations compared with FCMdd.
%its difference to other methods.
\end{abstract}
\begin{IEEEkeywords}
Credal partitions; Relational clustering; Evidential $c$-medoids; Imprecise classes.
\end{IEEEkeywords}

% \fi
\IEEEpeerreviewmaketitle

\section{Introduction}
Clustering  is a useful technique to detect the underlying cluster structure of the data set. The goal of clustering is to partition  a set of objects $X=\{x_1,x_2,\cdots,x_n\}$ into $c$ small subgroups $\Omega=\{\omega_1,\omega_2,\cdots,\omega_c\}$ based on a well defined measure of similarities between patterns. To measure the similarities (or dissimilarities), the objects are described by either object data or relational data. Object data are described explicitly by a feature vector, while relational data arise from the  pairwise similarities or dissimilarities. Among the existing approaches to clustering, the objective function-driven or
prototype-based clustering  such as $C$-Means (CM) and Fuzzy $C$-Means (FCM)  is
one of the most widely applied paradigms in statistical pattern recognition.
These methods are based on a fundamentally very simple, but nevertheless very
effective idea, namely to describe the data under consideration by a set of
prototypes. They  capture the characteristics of the data distribution (like
location, size, and shape), and classify the data set based on
the similarities (or dissimilarities) of the objects to their prototypes.

The above mentioned clustering algorithms, CM and  FCM are for object data. The prototype of each class in these methods is the center of gravity of all the included patterns.  But for relational data set, it is difficult to determine the centers of objects. In this case, one of the objects which is most similar to the center could be the most rational choice to be setting as the prototype. This is the idea of clustering using medoids.  Some clustering methods, such as Partitioning Around Medoids (PAM) \cite{kaufman2009finding} and  Fuzzy $C$-Medoids (FCMdd)  \cite{krishnapuram2001low}, produce hard and  soft clusters where each of them is represented by a representative object (medoid).
%However, in real applications, in order to capture various aspects of the data structures, we may need more members rather than one to be referred as the prototypes of each group.

Belief functions have already been  applied in many fields,
such as data classification \cite{liu2014credal}, data clustering \cite{masson2008ecm, liu2015credal}, social network analysis \cite{zhou2014evidential,zhou2015median} and statistical estimation \cite{denoeux2013maximum,zhou2014evidentialem}.  Evidential $C$-means (ECM) \cite{masson2008ecm} is a newly proposed clustering method to get credal partitions  for object data. The credal partition is  a general extension of the  crisp (hard) and fuzzy ones and it allows the object
to belong to not only  single clusters,  but also any subsets of the set of clusters
$\Omega=\{\omega_1,\cdots,\omega_c\}$  by allocating a mass of belief for each object in $X$ over the power
set $2^{\Omega}$. The additional flexibility brought by the power set
provides more refined partitioning results than those by the other  techniques
allowing us to gain a deeper insight into the data \cite{masson2008ecm}. In this paper, we introduce an extension of FCMdd on the framework of belief functions.  The evidential clustering algorithm for relational data sets,  named ECMdd, using a medoid which is assumed to belong to the original data set to represent a class are proposed to produce the optimal credal partition. The experimental results show the effectiveness of the methods and illustrate the advantages of credal partitions.

The rest of this paper is organized as follows. In Section II, some basic knowledge and the rationale of our method are briefly introduced. In Section
III  the proposed ECMdd clustering approach is presented in detail. In Section IV we test ECMdd using various data sets and  compare it  with several other classical methods. Finally, we conclude and present some perspectives in Section V.

\section{Background}
\subsection{Theory of belief functions}
Let $\Omega=\{\omega_{1},\omega_{2},\ldots,\omega_{c}\}$ be the finite domain of
$X$, called the discernment frame. The belief functions are defined on the power
set $2^{\Omega}=\{A:A\subseteq\Omega\}$.

%\begin{definition}
The function $m:2^{\Omega}\rightarrow[0,1]$ is said to be the Basic Belief
Assignment (bba) on $2^{\Omega}$, if it satisfies:
\begin{equation}
\sum_{A\subseteq\Omega}m(A)=1.
\end{equation}
Every $A\in2^{\Omega}$ such that $m(A)>0$ is called a focal element.
%\end{definition}
The credibility and plausibility functions are defined as in Eq.$~\eqref{bel}$ and Eq.$~\eqref{pl}$.
\begin{equation}
Bel\text{(}A\text{)}=\sum_{B\subseteq A, B \neq \emptyset} m\text{(}B\text{)} ~~\forall A\subseteq\Omega,
\label{bel}
\end{equation}

\begin{equation}
 Pl\text{(}A\text{)}=\sum_{B\cap A\neq\emptyset}m\text{(}B\text{)},~~\forall A\subseteq\Omega.
 \label{pl}
\end{equation}
Each quantity $Bel(A)$  measures the total support given to $A$, while $Pl(A)$ represents potential amount of support to $A$.
%Functions $Bel$ and $Pl$ are linked by the following relation:
%\begin{equation}
%  Pl(A) = 1 - m(\emptyset)-Bel(\overline{A}),
%\end{equation}
%where $\overline{A}$ denotes the complement of $A$ in $\Omega$.

A belief function on the credal level can be transformed into a probability function by Smets method \cite{smets2005decision}. In this
algorithm, each mass of belief $m(A)$ is
equally distributed among the elements of $A$.
This leads to the concept of pignistic probability, $BetP$, defined by
\begin{equation}
 \label{pig}
	BetP(\omega_i)=\sum_{\omega_i  \in A \subseteq \Omega } \frac{m(A)}{|A|(1-m(\emptyset))},
\end{equation}
where $|A|$ is the number of elements of $\Omega$ in $A$.
\subsection{Evidential $c$-means}
Evidential $c$-means \cite{masson2008ecm} is a direct generalization of FCM in the
framework of belief functions based on the concept of credal partitions.
%first %proposed in \cite{denoeux2004evclus}.
The credal partition takes advantage
of imprecise (meta) classes  to express partial knowledge of class memberships.  In ECM, the evidential membership of an object
$x_i$ is represented by a
bba $m_i=\left(m_i\left(A_k\right): A_k \subseteq \Omega \right)$ $(i=1,2,\cdots,n)$ over the given
frame of discernment $\Omega$. The set $\left\{A_k\mid A_k \subseteq \Omega, k=1,2,\cdots,2^c\right\}$ contains all the focal elements. The
optimal credal partition is obtained by  minimizing the following objective
function:
\begin{equation}
	J_{\mathrm{ECM}}=\sum\limits_{i=1}^{n}\sum\limits_{A_k\subseteq \Omega,A_k
	\neq \emptyset}|A_k|^\alpha
	m_{i}(A_k)^{\beta}d_{ik}^2+\sum\limits_{i=1}^{n}\delta^2m_{i}(\emptyset)^{\beta}
	\label{JECM}
\end{equation}
constrained on
\begin{equation}
	\sum\limits_{A_k\subseteq \Omega,A_k \neq
	\emptyset}m_{i}(A_k)+m_{i}(\emptyset)=1, \label{ECMconstraint}
\end{equation}
and
\begin{equation}
\label{nonegcon}
  m_{i}\left(A_k\right) \geq 0, ~~ m_{i}\left(\emptyset \right) \geq 0,
\end{equation}
where $m_{i}(A_k)\triangleq m_{ik}$ is the bba of $x_i$ given
to the nonempty set $A_k$, while $m_{i}(\emptyset)\triangleq m_{i\emptyset}$
is the bba of $x_i$ assigned to the empty set. Parameter $\alpha$ is a tuning parameter allowing to control the
degree of penalization for subsets with high cardinality, parameter $\beta$ is
a weighting exponent and  $\delta$ is an adjustable  threshold  for  detecting
the outliers.
%It is noted that for credal partitions, $k$ is not from 1 to $c$
%as usual, but ranges in $[0,f]$ with $f=2^c$.
Here $d_{ik}$ denotes the
distance (generally Euclidean distance) between $x_i$ and the barycenter ({\em i.e.} prototype,
denoted by $\overline{v}_k$) associated with $A_k$:
\begin{equation}\label{dis_node_pro}
  d_{ik}^2=\|x_i-\overline{v}_k\|^2,
\end{equation}
where $\overline{v}_k$ is defined mathematically by
\begin{equation}\label{prototypes}
	\overline{v}_k=\frac{1}{|A_k|}\sum_{h=1}^c s_{hk} v_h,
	~~\text{with}~~ s_{hk}=
  \begin{cases}
        1 & \text{if} ~~ \omega_h \in A_k \\
        0 & \text{else}\end{cases}.
 \end{equation}
The notation $v_h$ is the geometrical
center of points in cluster $h$. The update process with Euclidean distance  is given by the
following two alternating steps.
%\iffalse
\begin{itemize}
\item Assignment update, $\forall i, ~\forall k/A_k  \subseteq  \Omega, A_k \neq \emptyset$:
\begin{align}
\label{ECM_updata_bba}
m_{ik}=\frac{|A_k|^{-\alpha/(\beta-1)}{d_{ik}^{-2/(\beta-1)}}}{\sum\limits_{A_h\neq\emptyset}
|A_h|^{-\alpha/(\beta-1)}{d_{ih}^{-2/(\beta-1)}}+\delta^{-2/(\beta-1)}},
\end{align}
and for $A_k = \emptyset$
 \begin{align}
			m_{i\emptyset}=1-\sum_{A_k\neq \emptyset}m_{ik}, ~~\forall
		i=1,2,\cdots,n.
\end{align}
\item Prototype update: The prototypes
		(centers) of the classes are given by the rows of the matrix
		$v_{c\times p}$, which is the solution of the following linear system:
	\begin{equation}
\label{HB}
    \bm{HV}=\bm{B},
   \end{equation}
where $\bm{H}$ is a matrix of size $(c \times c)$ given by
\begin{equation}\label{H}
   \bm{H}_{lk}=\sum_i
			\sum_{A_k \supseteqq \{\omega_k,\omega_l\}} |A_k|^{\alpha-2}
			m_{ik}^\beta,
\end{equation} and $\bm{B}$ is a matrix of size $(c \times p)$ defined by
\begin{equation}\label{B}
			\bm{B}_{lq}=\sum_{i=1}^n x_{iq}\sum_{A_k\ni
			\omega_l}|A_k|^{\alpha-1}m_{ik}^\beta.
\end{equation} \end{itemize}

\subsection{Fuzzy $c$-medoids}
Fuzzy $C$-Medoids (FCMdd) is a variation of classical $c$-means clustering designed for relational data \cite{krishnapuram2001low}. Let \linebreak $\bm{X}=\left\{x_i\mid i=1,2,\cdots,n\right\}$ be the set of $n$ objects and $\tau(x_i, x_j)\triangleq \tau_{ij}$ denote the dissimilarity between objects $x_i$ and  $x_j$. Each object may or may not be represented by a feature vector. Let $\bm{V}= \{v_1, v_2,\cdots,v_c\}$, $v_i \in \bm{X}$ represent a subset of $\bm{X}$. The objective function of FCMdd is given as
\begin{equation}
  J_\text{FCMdd} =  \sum_{i=1}^n \sum_{j=1}^c u_{ij}^\beta \tau(x_i, v_j)
\end{equation}
subject to
\begin{equation}
  \sum_{j=1}^c u_{ij}=1, i=1,2,\cdots,n,
\end{equation}
and
\begin{equation}
  u_{ij}\geq 0, i=1,2,\cdots,n, ~~j=1,2,\cdots,c.
\end{equation}
In fact, the objective function of FCMdd is similar to that of FCM. The main difference lies in that the prototype of a class in FCMdd is defined as the medoid, {\em i.e.,} one of the object in the original data set, instead of the centroid (the average point in a continues space) for FCM. FCMdd is
preformed by the following alternating update steps:
\begin{itemize}
\item
		Assignment update:
\begin{equation}
		u_{ij}=\frac{\tau_{ij}^{-1/(\beta-1)}}{\sum\limits_{k=1}^c
	\tau_{ik}^{-1/(\beta-1)}}.
\end{equation}
\item Prototype update: the new
	prototype of cluster $j$ is set to be $v_{j}=x_{l^*}$ with
	\begin{equation} x_{l^*}= \arg \min_{\{v_j:v_j=x_l
	(\in X)\}} \sum_{i=1}^n u_{ij}^\beta \tau(x_i,v_j).
 \end{equation}
\end{itemize}

\section{Evidential $c$-medoids clustering}
Here we introduce evidential $c$-medoids clustering algorithm using medoids in order to take advantages of both medoid-based clustering and credal partitions. This partitioning evidential clustering algorithm is mainly related to fuzzy $c$-medoids.  Like all the prototype-based clustering methods, for ECMdd,
an objective function should first be  found to provide an immediate measure of
the quality of partitions. Hence our goal can  be characterized as
the optimization of the objective function to get the best credal partition.
\subsection{The objective function}
As before, let $\bm{X}=\left\{x_i\mid i=1,2,\cdots,n\right\}$ be the set of $n$ objects and $\tau(x_i, x_j)\triangleq \tau_{ij}$ denote the dissimilarity between objects $x_i$ and  $x_j$. The pairwise dissimilarity is the only information required for the analyzed data set.
%Each object may or may not be represented by a feature vector.
The objective function of ECMdd is similar to that in ECM:
\begin{equation} J_{\mathrm{ECMdd}}(\bm{M},
	\bm{V})=\sum\limits_{i=1}^{n}\sum\limits_{A_j\subseteq \Omega,A_j \neq
	\emptyset}|A_j|^\alpha
	m_{ij}^{\beta}d_{ij}+\sum\limits_{i=1}^{n}\delta^2m_{i\emptyset}^{\beta},
	\label{costfun} \end{equation}
\noindent constrained on
\begin{equation}
	\sum\limits_{A_j\subseteq \Omega,A_j \neq
	\emptyset}m_{ij}+m_{i\emptyset}=1, \label{ECMddconstraint}
\end{equation}
where $m_{ij}\triangleq m_{i}(A_j)$ is the bba of $x_i$ given to the nonempty
set $A_j$, $m_{i\emptyset} \triangleq m_{i}(\emptyset)$ is the bba of $x_i$
assigned to the empty set, and $d_{ij}\triangleq d(x_i, A_j)$ is the dissimilarity
between $x_i$ and  focal set $A_j$. Parameters $\alpha,\beta,\delta$ are adjustable with the same meanings
as those in ECM. Note that $J_\text{ECMdd}$ depends on the credal partition $\bm{M}$ and the set $\bm{V}$ of all
prototypes.

Let  $v_k^\Omega$ be the prototype of  specific  cluster (whose focal element is a singleton) $A_j=\{\omega_k\}$ $(k=1,2,\cdots,c)$ and assume that it must be one of the objects in $X$. The dissimilarity between object $x_i$ and cluster (focal set) $A_j$  can be defined as follows. If $\left|A_j\right|=1$, {\em i.e.}, $A_j$ is associated with one of the
singleton clusters in $\Omega$ (suppose to be $\omega_k$ with prototype
$v_k^\Omega$, {\em i.e.,} $A_j=\{\omega_k\}$), then the dissimilarity between $x_i$ and $A_j$ is defined by
\begin{equation} \label{specific_pro}
	d_{ij}=d(x_i, A_j) = \tau(x_i,v_k^\Omega). \end{equation}
When $|A_j|>1$, it represents an imprecise (meta)
cluster. If object $x_i$ is to be partitioned into a meta cluster, two
conditions should be satisfied \cite{zhou2015median}. One condition is  the dissimilarity values between
$x_i$ and the included singleton classes'  prototypes are small. The
other condition is the object should be close to   the prototypes of all these  specific
clusters.  The former measures the degree of  uncertainty, while the latter is
to avoid the pitfall of  partitioning two data objects irrelevant to any
included specific clusters into the corresponding imprecise classes.  Therefore, the medoid (prototype) of an imprecise class $A_j$ could be set to be one of the objects locating with similar dissimilarities to all the prototypes of the specific classes $\omega_k \in A_j$ included in $A_j$. The variance of the dissimilarities of object $x_i$ to the medoids of all the included specific classes of $A_j$ could be taken into account to express the degree of uncertainty. The smaller the variance is, the higher uncertainty we have for object $x_i$. Meanwhile the medoid should be close to all the prototypes of the specific
classes.  This is to distinguish the outliers, which may have equal dissimilarities to the prototypes of some specific classes, but obviously not a good choice for representing the associated imprecise classes. Let $v^{2^\Omega}_j$ denote the medoid of class $A_j$\footnote{The notation $v^\Omega_k$ denotes the prototype of specific class $\omega_k$, thus it is in the framework of $\Omega$. Similarly, $v^{2^\Omega}_j$ is defined on the power set $2^\Omega$, representing the prototype of the focal set $A_j$ $\in 2^\Omega$. It is easy to see $\{v_k^{\Omega}:k=1,2,\cdots,c\}\subseteq \{v_j^{2^\Omega}:j=1,2,\cdots,2^c-1\}$.}. Based on the above analysis, the medoid of $A_j$ should set to $v_j^{2^\Omega}=x_p$ with
\begin{align}
\label{imepro}
  p = \arg \min\limits_{i:x_i\in X} \Big\{ f\left(\{\tau(x_i, v_k^\Omega); \omega_k \in A_j \}\right) \nonumber \\ + \eta \frac{1}{|A_j|} \sum\limits_{\omega_k \in A_j} \tau(x_i, v_k^\Omega)\Big\},
\end{align}
where $\omega_k$ is the element of $A_j$, $v_k^\Omega$ is its corresponding prototype and $f$ denotes the function describing the variance among the corresponding dissimilarity values.
The  variance function could be used directly:
\begin{equation}
  \text{Var}_{ij}= \frac{1}{|A_j|}\sum_{\omega_k \in A_j} \bigg[\tau(x_i, v_k^{\Omega})-\frac{1}{|A_j|}\sum_{\omega_k \in A_j}\tau(x_i, v_k^{\Omega})\bigg]^2.
\end{equation}
In this paper, we use the following function to describe  the variance $\rho_{ij}$ of the dissimilarities between object $x_i$ and the medoids of the involved specific classes in $A_j$ :
\begin{equation}
  \rho_{ij}= \frac{1}{\text{choose}(|A_j|,2)}\sum\limits_{\omega_x, \omega_y \in A_j} \sqrt{\left(\tau(x_i, v_x^{\Omega})-\tau(x_i, v_y^{\Omega})\right)^2},
\end{equation}
where  $\text{choose}(a,b)$ is the number of combinations of the given $a$ elements taken $b$ at a time.

The dissimilarity between objects $x_i$ and class $A_j$ can be defined as
\begin{equation}
  d_{ij} = \frac{\tau(x_i, v^{2^\Omega}_j) + \gamma \frac{1}{|A_j|} \sum\limits_{\omega_k \in A_j} \tau(x_i, v_k^\Omega)}{1+\gamma}.
\end{equation}
As we can see from the above equation, the dissimilarity between object $x_i$ and meta class $A_j$ ($|A_j|>1$) is the weighted average of dissimilarities of $x_i$ to the all involved  singleton cluster medoids and to the prototype of the imprecise class  $A_j$ with a tuning factor $\gamma$. If $A_j$ is a specific class with $A_j=\{\omega_k\}$ ($|A_j|=1$), the dissimilarity between $x_j$ and $A_j$ degrades to the dissimilarity between $x_i$ and $v_k^{\Omega}$ as defined in Eq.~\eqref{specific_pro}, {\em i.e.,} $v^{2^\Omega}_j=v_k^\Omega$. And if $|A_j|>1$, its medoid is decided by Eq.~\eqref{imepro}.

%\noindent\textbf{Remark 1:}
It is remarkable that although ECMdd is similar to Median Evidential $C$-Means (MECM) \cite{zhou2015median} algorithm in principle, but they are very different in dealing with the imprecise classes % MECM is in the framework of median clustering, while ECMdd consists with FCMdd in principle.
%Another difference of ECMdd and MECM is
and the way of calculating the dissimilarities between objects and imprecise classes.   Although both MECM and ECMdd consider the dissimilarities of objects to the prototypes for specific clusters, the strategy adopted by ECMdd is more simple and  intuitive. Moreover, there is no representative medoid for imprecise classes in MECM.
\subsection{The optimization}
To minimize $J_\text{ECMdd}$, an optimization
scheme via an Expectation-Maximization (EM) algorithm  can be
designed, and the alternate update steps are as follows:

\noindent Step 1. Credal partition ($\bm{M}$) update.

The bbas of objects' class membership for any subset \linebreak $A_j \subseteq \Omega$ and the empty set $\emptyset$ representing the outliers  are updated identically to ECM \cite{masson2008ecm}:
\begin{itemize} \item
			$\forall A_j \subseteq \Omega, A_j \neq \emptyset$,
			\begin{equation} \label{mass1}
m_{ij}=\frac{|A_j|^{-\alpha/(\beta-1)}{d_{ij}^{-1/(\beta-1)}}}
{\sum\limits_{A_k\neq\emptyset}|A_k|^{-\alpha/(\beta-1)}{d_{ik}^{-1/(\beta-1)}}+\delta^{-1/(\beta-1)}}
		\end{equation}
\item If $A_j = \emptyset$, \begin{equation}
				\label{mass2} m_{i\emptyset}=1-\sum\limits_{A_j \neq
				\emptyset}m_{ij} \end{equation} \end{itemize} \noindent Step
2. Prototype ($\bm{V}$) update.

The prototype $v_i^\Omega$ of a specific (singleton) cluster \linebreak $\omega_i$
$(i=1,2,\cdots,c)$ can be updated first and then the prototypes of imprecise (meta) classes could be determined by Eq.~\eqref{imepro}.  For
singleton clusters $\omega_k$ $(k=1,2,\cdots,c)$, the corresponding new
prototype $v_k^\Omega$ $(k=1,2,\cdots,c)$  could be set to $x_l \in X$ such that
  \begin{equation}
\label{pro_update}
 x_l= \arg \min_{v^{'}_k}
	\left\{\sum_{i=1}^n  \sum_{A_j = \{\omega_k\}}
	 m_{ij}^\beta d_{ij}(v^{'}_k): v^{'}_k
	\in X\right\}.
\end{equation}
The dissimilarity between object $x_i$ and cluster $A_j$, $d_{ij}$, is a function of $v^{'}_k$, which is the potential prototype of class $\omega_k$.
%\end{enumerate}

The bbas of the objects' class assignment are updated  identically to ECM
\cite{masson2008ecm}, but it is worth noting that $d_{ij}$ has
different meanings as that in ECM although in both cases it measures the dissimilarity between object $x_i$ and class $A_j$. In ECM $d_{ij}$ is the distance between object $i$ and the centroid point of $A_j$, while in ECMdd, it is the dissimilarity between $x_i$ and the most ``possible" medoid.  For the
prototype updating process the fact that the prototypes are assumed to be one
of the data objects is taken into consideration. Therefore, when the credal
partition matrix $\bm{M}$ is fixed, the new prototype of each cluster can be
obtained in a simpler manner than in the case of ECM application. The ECMdd algorithm is summarized as Algorithm \ref{alg:method}.

We discuss here about the convergence of ECMdd. The assignment update process will not increase $J_\text{ECMdd}$  since the new mass matrix is determined by differentiating of
the respective Lagrangian of the cost function with respect to $\bm{M}$. Also $J_\text{ECMdd}$ will not increase through the medoid-searching scheme for prototypes of specific classes. If the prototypes of specific classes are fixed, the medoids of imprecise classes determined by Eq.~\eqref{imepro} are likely to locate near to the ``centroid" of all the prototypes of the included specific classes. If the objects are in Euclidean space, the medoids of imprecise classes are near to the centroids found in ECM. Thus it  will not increase the value of the objective function also. Moreover,
the bba $\bm{M}$ is a function of the prototypes $\bm{V}$ and for given $\bm{V}$ the
assignment $\bm{M}$ is unique. Because ECMdd assumes that the prototypes are
original object data in  $\bm{X}$, so there is a finite number of
different prototype vectors $\bm{V}$ and so is the number of corresponding credal
partitions $\bm{M}$. Consequently we can conclude that the ECMdd algorithm
converges in a finite number of steps.

\begin{algorithm}\caption{\textbf{:} ~~~ECMdd  algorithm}\label{alg:method}
\begin{algorithmic}
\STATE {\textbf{Input:} Dissimilarity matrix $[\tau(x_i,x_j)]_{n\times n}$ for the $n$ objects
	$\{x_1,x_2,\cdots,x_n\}$.}
\STATE{\textbf{Parameters:}
	~\\$c$: number clusters $1<c<n$ \\ $\alpha$:
	weighing exponent for cardinality \\ $\beta >1$: weighting
	exponent \\ $\delta>0$: dissimilarity between any object to
	the empty set \\ $\eta>0$: to distinguish the outliers from the possible medoids\\
$\gamma \in [0,1]$: balance of the contribution for imprecise classes\\}
\STATE {\textbf{Initialization:}\\ Choose
			   randomly $c$ initial prototypes from the object set
			}
\REPEAT
\STATE{
 (1). $t\leftarrow t+1$\\ (2). Compute $\bm{M}_t$ using
Eq.~\eqref{mass1}, Eq.~\eqref{mass2} and  $\bm{V}_{t-1}$\\ (3).
Compute the new prototype set $\bm{V}_{t}$ using Eq.~\eqref{pro_update} and \eqref{imepro}
}
\UNTIL{the prototypes remain unchanged.}
\STATE {\textbf{Output:} The optimal credal partition.}
\end{algorithmic}
\end{algorithm}

%\noindent \textbf{Remark 2:}

\subsection{The parameters of the algorithm}
As in ECM, before running ECMdd,
the values of the parameters have to be set. Parameters $\alpha, \beta$ and
$\delta$ have the same meanings as those in ECM. The value $\beta$ can be set to be $\beta=2$ in all experiments for which it
is a usual choice. The parameter $\alpha$  aims to penalize the subsets with
high cardinality and control  the amount of points assigned to imprecise
clusters for credal partitions.  The higher
$\alpha$ is, the less mass belief is assigned to the meta clusters and the
less imprecise will be the resulting partition. However, the decrease of
imprecision may result in high risk of errors. For instance, in the case of hard
partitions, the clustering results are completely precise but there is much more intendancy  to partition an object to
an unrelated group. As suggested in \cite{masson2008ecm}, a value can be used as a
starting default one but it can be modified according to what is expected from
the user.  The choice $\delta$ is more difficult and is strongly data
dependent \cite{masson2008ecm}. In ECMdd, parameter $\gamma$ weighs the contribution of  uncertainty to the dissimilarity between objects and imprecise clusters. %If the analyzed data set is highly overlapped, $\gamma$ could be set relatively small to make the imprecise classes express more uncertainty.
Parameter $\eta$ is used to distinguish the outliers from the possible medoids when determining the prototypes of meta classes. It could be set 1 by default and it has little effect on the final partition results.

For determining the number of clusters, the validity index  of a credal
partition  defined by  \cite{masson2008ecm}
could be utilised:
\begin{align}\label{clu_num_ind} N^*(c)\triangleq \frac{1}{n \log_2(c)}
	\times \sum_{i=1}^n \Bigg[&\sum_{A\in 2^\Omega \setminus \emptyset}
	m_i(A)\log_2|A| \nonumber \\ &+m_i(\emptyset)\log_2(c)\Bigg], \end{align}
where $0 \leq N^*(c) \leq 1$. This index has to be minimized to get the optimal number
of clusters.

\section{Experiments}
In this section some experiments on various data sets will be performed to show the effectiveness of ECMdd. The results are compared with FCMdd and MECM to illustrate the effectiveness and merits of the proposed method.

The $c$-means type clustering algorithms are sensitive to the initial  prototypes. In this work, we follow the initialization procedure as the one used in \cite{krishnapuram2001low} and \cite{mei2010fuzzy} to generate a set of $c$ initial prototypes one by one. The first medoid, $\sigma_1$,  is randomly picked from the data set.
The rest of medoids are selected successively one by one in such a way that each one is most dissimilar to all the medoids that have already been picked. Suppose $\sigma=\{\sigma_1, \sigma_2, \cdots, \sigma_j\}$ is the  set of the first chosen $j$ ($j<c$) medoids. Then the $j+1$ medoid, $\sigma_{j+1}$, is set to the object $x_p$ with
\begin{equation}
 p = \arg\max\limits_{1\leq i \leq n; x_i \notin \sigma} \left\{\min\limits_{\sigma_k \in \sigma} \tau(x_i, \sigma_k)\right\}.
\end{equation}
This selection process makes the initial prototypes evenly distributed and locate as far away from each other as possible. The  popular  measures, Precision (P), Recall (R) and Rand Index (RI),   which  are  typically  used  to  evaluate the
performance  of hard clusterings are also used here. Precision  is the fraction  of relevant
instances (pairs in identical groups in  the clustering benchmark)  out  of
those  retrieved  instances (pairs in identical groups of the discovered
clusters), while recall is the fraction of relevant instances that are
retrieved. Then precision and recall can be calculated by
 \begin{equation}
	\label{precision} \text{P}=\frac{a}{a+c}
	~~~~~\text{and}~~~~~\text{R}=\frac{a}{a+d}
 \end{equation} respectively, where $a$ (respectively, $b$) be the number of pairs of
objects simultaneously assigned to identical classes (respectively, different
classes) by the stand reference partition and the obtained one. Similarly, values $c$ and $d$ are the numbers of dissimilar pairs  partitioned into the same cluster, and the number of similar object pairs clustered into
different clusters respectively.
The rand index  measures the percentage of correct decisions and it can be defined as
\begin{equation} \label{ri}
	\text{RI}=%\frac{\text{TP}+\text{TN}}{\text{TP}+\text{TN}+\text{FP}+\text{FN}}=
	\frac{2(a+b)}{n(n-1)},
 \end{equation}
where $n$ is the number of data objects.

For fuzzy and evidential clusterings,  objects  may  be partitioned into
multiple  clusters with different degrees.  In  such cases precision  would
be  consequently  low \cite{mendes2003evaluating}.
Usually the  fuzzy and
evidential clusters  are  made  crisp  before calculating  the  measures,
using  for  instance  the  maximum membership criterion
\cite{mendes2003evaluating} and pignistic probabilities
\cite{masson2008ecm}. Thus in this work we will
harden the fuzzy and credal clusters by maximizing the corresponding
membership and pignistic probabilities and calculate precision, recall and RI
for each case.

The introduced imprecise clusters can avoid the risk to group a data into a
specific class without strong belief. In other words, a data pair can be
clustered into the same specific group only when we are quite confident and
thus the misclassification rate will be reduced. However, partitioning too
many data into  imprecise clusters may cause that  many objects are not
identified for their precise groups. In order to show the effectiveness of the
proposed method in these aspects, we use the indices for evaluating credal partitions, Evidential Precision (EP),
Evidential Recall (ER) and Evidential Rank Index (ERI) \cite{zhou2015median} defined as:
\begin{equation}\label{ep}
	\text{EP}=\frac{n_{er}}{N_e}, ~~~ \text{ER}=\frac{n_{er}}{N_r}, ~~~\text{ERI}=\frac{2(a^*+b^*)}{n(n-1)}.
\end{equation}
In Eq.~\eqref{ep}, the notation $N_e$ denotes the number of
pairs partitioned into the same specific group by evidential clusterings, and
$n_{er}$ is the number of  relevant instance pairs out of these specifically
clustered pairs. The value $N_r$  denotes the number of pairs in the same
group of the clustering benchmark, and ER is the fraction of  specifically
retrieved instances (grouped into an identical specific cluster) out of these
relevant pairs. Value $a^*$ (respectively, $b^*$) is the number of pairs of
objects simultaneously clustered to the same specific class ({\em i.e.},
singleton class, respectively, different classes) by the stand reference
partition and the obtained credal one. When the partition degrades to a crisp one, EP, ER and ERI equal to
the classical precision, recall and rand index measures respectively.  EP and ER reflect
the accuracy of the credal partition from different points of view, but we
could not evaluate the clusterings from one single term. For example, if all
the objects are partitioned into imprecise clusters except two relevant data
object grouped into a specific class, $\text{EP}=1$ in this case. But we could
not say this is a good partition since it does not provide us with any information
of great value. In this case $\text{ER}\approx 0$. Thus ER could be used to express
the efficiency of the method for providing valuable partitions. ERI is like the combination of EP and ER
describing the accuracy of the clustering results.  Note that for  evidential  clusterings, precision, recall and RI  measures are
calculated after the corresponding hard partitions are got, while EP, ER and
ERI are based on hard credal partitions \cite{masson2008ecm}.

\subsection{Karate Club network}
\begin{center} \begin{figure}[!thbt] \centering
		\includegraphics[width=0.45\linewidth]{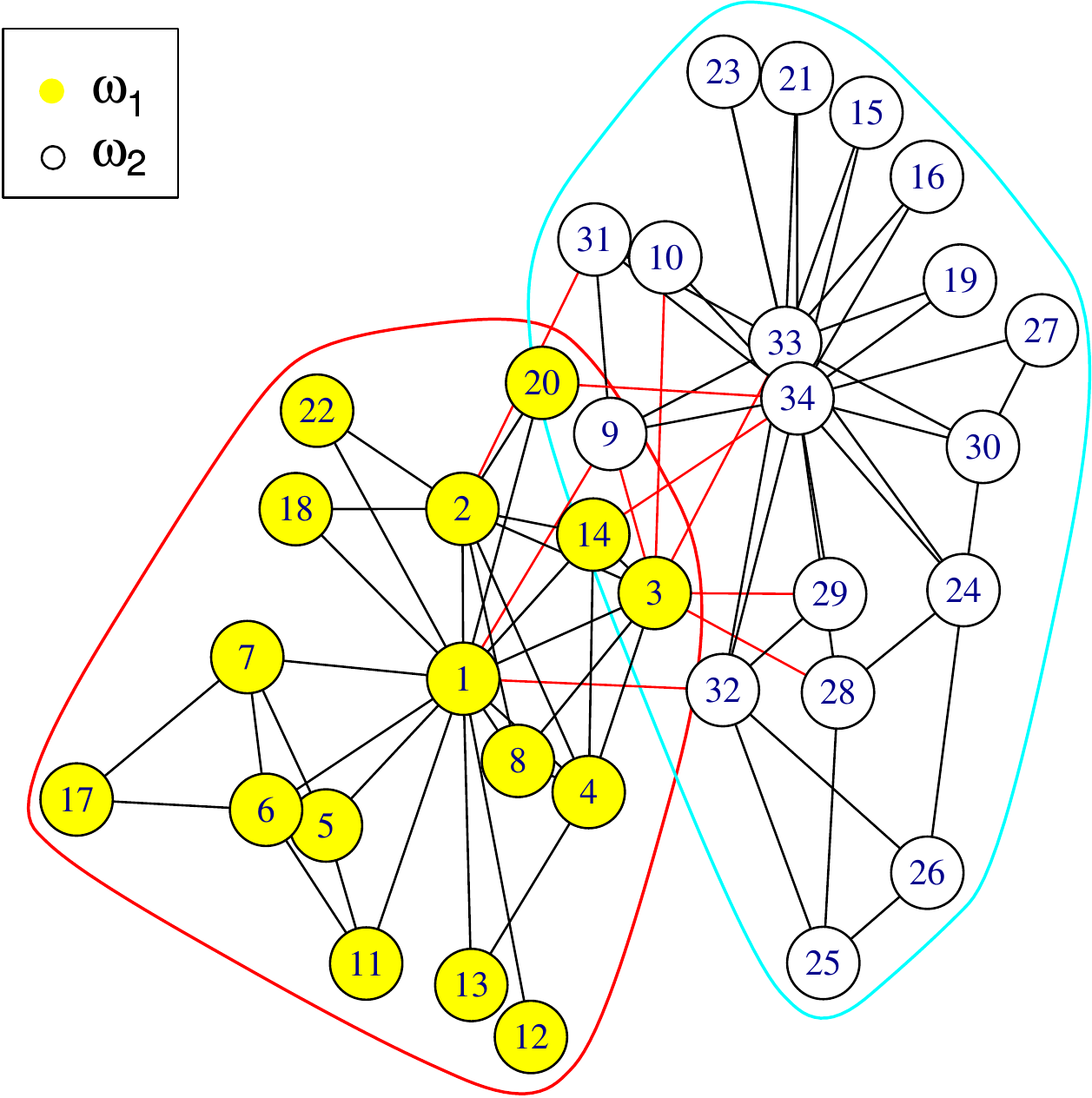}\hfill
        \includegraphics[width=0.45\linewidth]{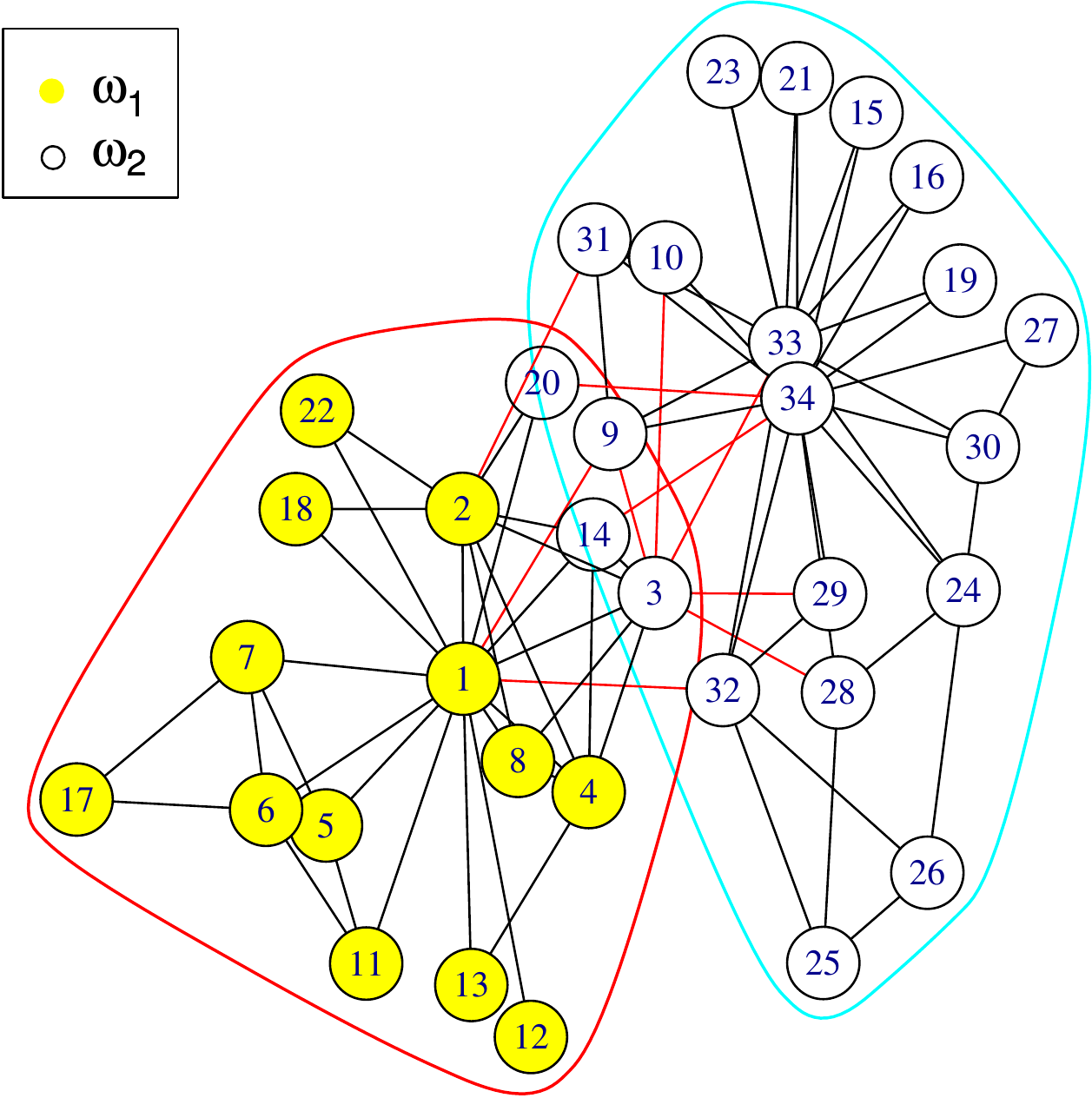} \hfill
        \parbox{.45\linewidth}{\centering\small a. Original network} \hfill
		\parbox{.45\linewidth}{\centering\small b. Results by FCMdd}
		\includegraphics[width=0.45\linewidth]{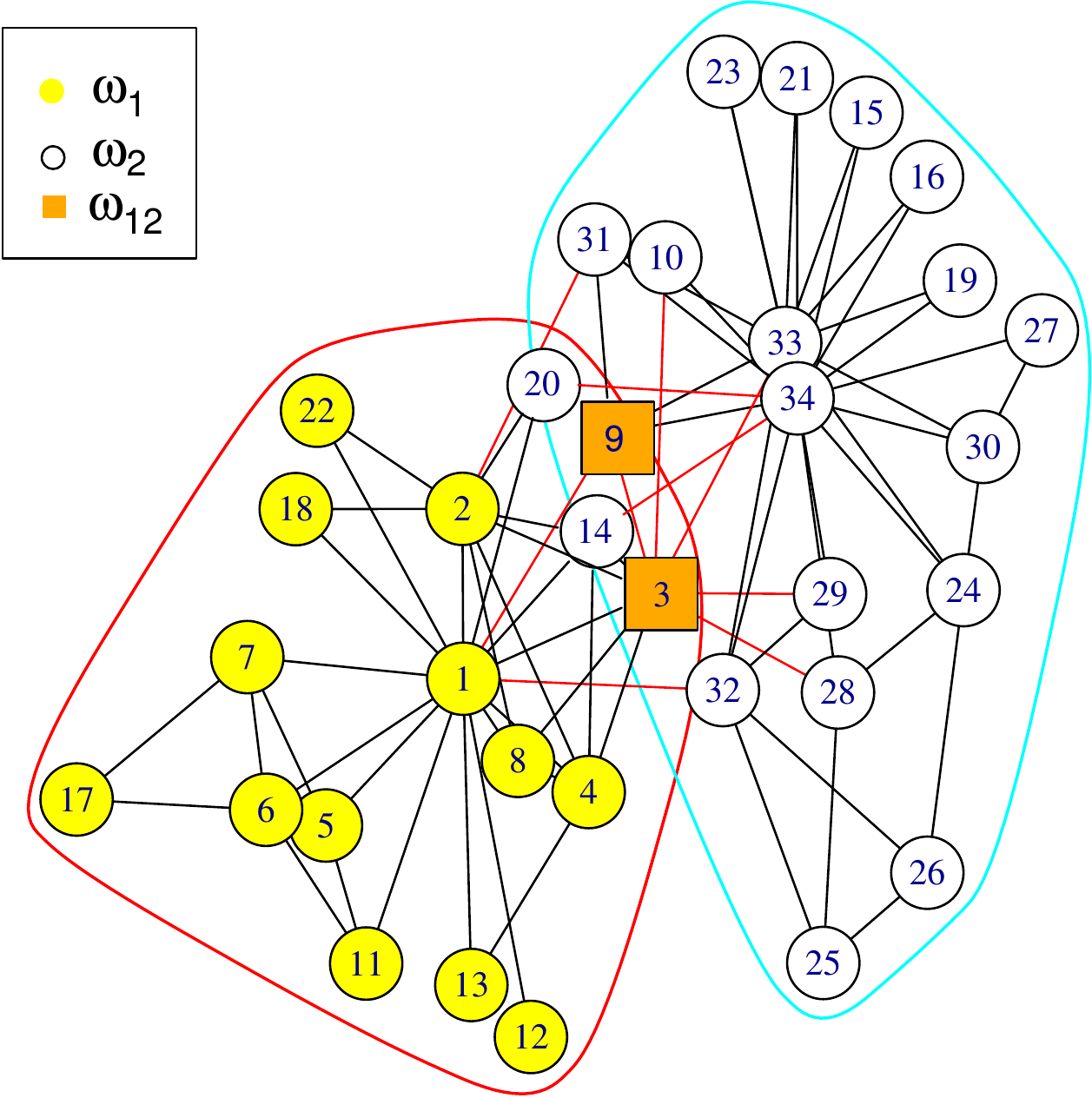}\hfill
        \includegraphics[width=0.45\linewidth]{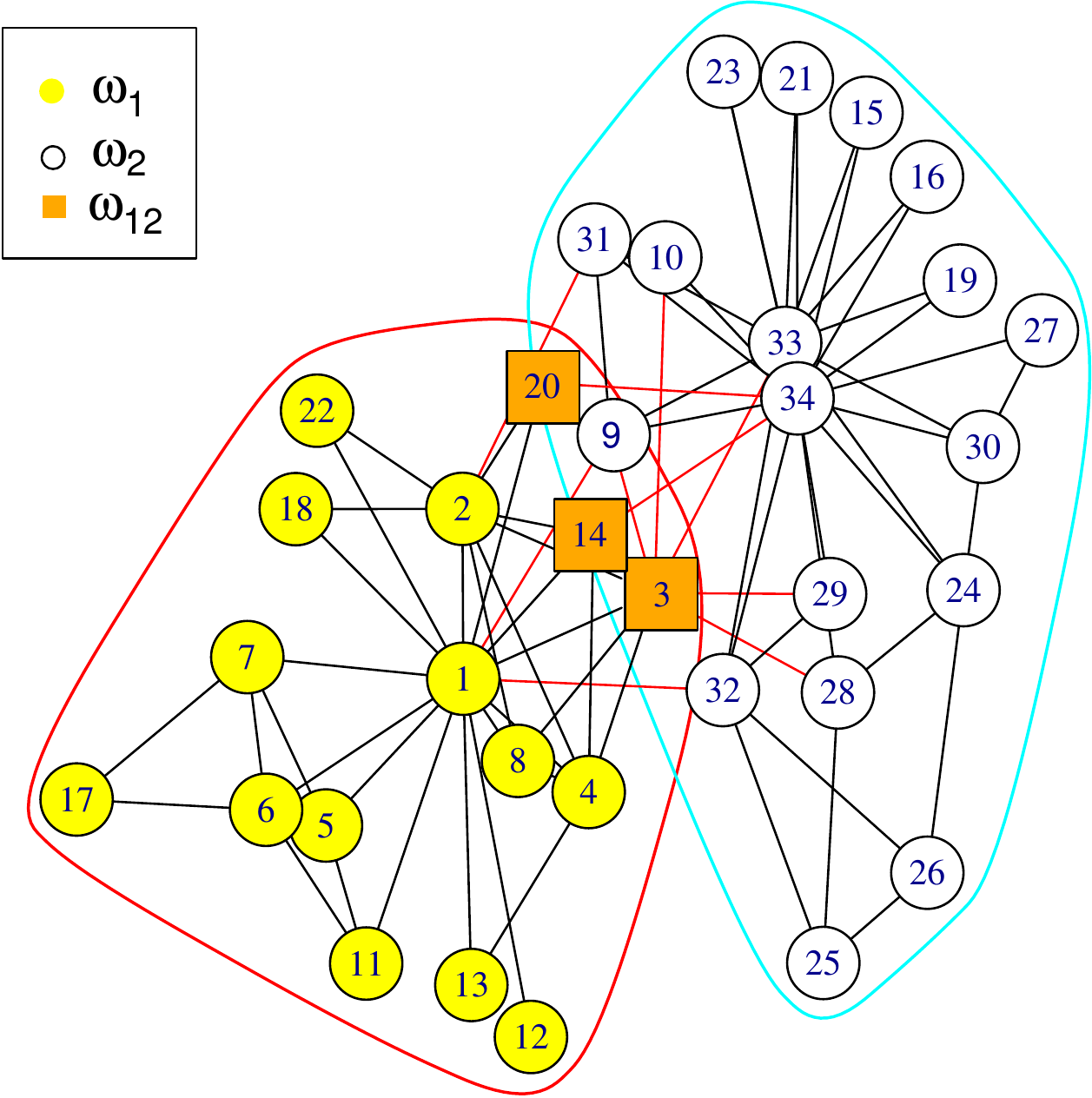} \hfill
        \parbox{.45\linewidth}{\centering\small c. Results by MECM} \hfill
		\parbox{.45\linewidth}{\centering\small d. Results by ECMdd}
\caption{The Karate Club network. The parameters of MECM are $\alpha=1.5, \beta=2, \delta=100, \eta=0.9, \gamma=0.05$. In ECMdd, $\alpha=0.05, \beta=2, \delta = 100, \eta = 1, \gamma =1$, while in FCMdd, $\beta=2$.}
\label{karate} \end{figure} \end{center}

Graph visualization is commonly used to visually model relations in many areas. For graphs such as social networks, the
prototype  of one group is likely to be one of the persons ({\em i.e.,}
nodes in the graph) playing the leader role in the community. Moreover, a graph (network) of vertices  and edges usually describes the interactions between different agents of the complex system and the pair-wise relationships between nodes are often implied in the graph data sets. Thus medoids-based relational clustering algorithms could be directly applied. In this section we will evaluate the effectiveness of the proposed methods applied on community detection problems. Here we  test on a widely used benchmark  in detecting community structures, ``Karate Club",  studied by Wayne Zachary. The network consists of 34  nodes and 78 edges representing the friendship among the members of the club (see Figure \ref{karate}.a). %During the development, a dispute arose between the club's administrator and instructor, which eventually resulted in the club  split into two smaller clubs, centered around the administrator and the instructor respectively.

There are many similarity and dissimilarity indices for networks, using local or global information of graph structure. In this experiment, different similarity metrics will be compared first. The similarity indices considered here are listed in Table \ref{indices}. It is notable that the similarities by these measures are from 0 to 1, thus they  could be converted  into dissimilarities simply by
$
  \textit{dissimilarity} = 1 - \textit{similarity}
$. The comparison results for different dissimilarity indices by FCMdd and ECMdd are shown in Table \ref{ComIndicesFCMdd} and Table \ref{ComIndicesECMdd} respectively.  As we can see, for all the dissimilarity indices, for ECMdd, the value of evidential precision is higher than that of precision. This can be attributed to the introduced imprecise classes which enable us not to make a hard decision for the nodes that we are uncertain and consequently guarantee the accuracy of the specific clustering results. From the table we can also see that the performance using the dissimilarity measure based on signal prorogation  is better than those using local similarities in the application of both FCMdd and ECMdd. This reflects that global dissimilarity metric is better than the local ones for community detection. Thus in the following experiments, we only consider the signal dissimilarity index.
\begin{table}[ht]
\centering \caption{Different local and global similarity indices.}
%\resizebox{\textwidth}{!}{
\begin{tabular}{rrrrrrr}
  \hline
 Index & Global metric& Ref.&Index  & Global metric& Ref.\\
  \hline
Jaccard & No &  \cite{jaccard1912distribution} &Zhou & No &  \cite{zhou2009predicting}\\
Pan  & No &  \cite{pan2010detecting} &Signal & Yes &\cite{hu2008community}\\
%Football network & \\
   \hline
\end{tabular}%}
\label{indices}
\end{table}
\vspace{-1.8em}
\begin{table}[ht]
\centering \caption{Comparison of different similarity indices by FCMdd.}
\begin{tabular}{rrrrrrr}
  \hline
 Index & P & R & RI & EP & ER & ERI \\
  \hline
Jaccard & 0.6364 & 0.7179 & 0.6631 & 0.6364 & 0.7179 & 0.6631 \\
  Pan & 0.4866 & 1.0000 & 0.4866 & 0.4866 & 1.0000 & 0.4866 \\
  Zhou & 0.4866 & 1.0000 & 0.4866 & 0.4866 & 1.0000 & 0.4866 \\
  Signal & 0.8125 & 0.8571 & 0.8342 & 0.8125 & 0.8571 & 0.8342 \\
   \hline
\end{tabular}\label{ComIndicesFCMdd}
\end{table}

\vspace{-1.8em}
% latex table generated in R 3.1.1 by xtable 1.7-3 package
% Sun Mar 08 12:11:22 2015
\begin{table}[ht]
\centering \caption{Comparison of different similarity indices by ECMdd.}
\begin{tabular}{rrrrrrr}
  \hline
 Index & P & R & RI & EP & ER & ERI \\
  \hline
Jaccard & 0.6458 & 0.6813 & 0.6631 & 0.7277 & 0.5092 & 0.6684  \\
Pan & 0.6868 & 0.7070 & 0.7005 & 0.7214 & 0.6923 & 0.7201 \\
Zhou  & 0.6522 & 0.6593 & 0.6631 & 0.7460 & 0.3443 & 0.6239\\
Signal & 1.0000 & 1.0000 & 1.0000 & 1.0000 & 0.6190 & 0.8146 \\
   \hline
\end{tabular}\label{ComIndicesECMdd}
\end{table}

The detected community structures by different methods are displayed in Figure \ref{karate}.b -- \ref{karate}.d.  FCMdd could detect the exact community structure of all the nodes except nodes 3, 14, 20. As we can see from the figures, these three nodes have connections  with both communities.  They are partitioned into imprecise class $\omega_{12}\triangleq\{\omega_1, \omega_2\}$, which describing the uncertainty on the exact class labels of the three nodes, by the application of ECMdd.   The medoids found by FCMdd of the two specific communities are node 5 and node 29, while by ECMdd node 5 and node 33.  The uncertain nodes found by MECM are node 3 and node 9.

From this experiment we can see that the introduced imprecise classes by credal partitions could help us make soft decisions for the uncertain objects which may lie in the overlapped area. This could avoid the risk of making errors simply by hard partitions.

% latex table generated in R 3.1.1 by xtable 1.7-3 package
% Mon Feb 23 11:10:25 2015
%\begin{table*}[ht]
%\centering \caption{The prototype weights by wECMdd-0.}
%\begin{tabular}{rrrrrrrrrrrrrrrrrrrrrrrrrrrrrrrrrrr}
%  \hline
%&  & 1 & 2 & 3 & 4 & 5 & 6 & 7 & 8 & 9 & 10 \\
%  \hline
%Community $\omega_1$& Node & 12 & 1 & 18 & 22 & 13 & 4 & 8 & 2 & 5 & 11 \\
% & Weights & 0.1639 & 0.1491 & 0.1054 & 0.1054 & 0.1043 & 0.0691 & 0.0674 & 0.0651 & 0.0339 & 0.0339 \\
%  \hline
% Community $\omega_2$& Node & 24 & 33 & 30 & 15 & 16 & 19 & 21 & 23 & 34 & 27 \\
%  &Weights & 0.1131 & 0.0931 & 0.0870 & 0.0855 & 0.0855 & 0.0855 & 0.0855 & 0.0855 & 0.0653 & 0.0514 \\
%  \hline
%  Community $\omega_{12}$&Node & 9 & 10 & 32 & 29 & 20 & 3 & 31 & 25 & 14 & 28 \\
%  & Weights & 0.9943 & 0.0016 & 0.0013 & 0.0009 & 0.0008 & 0.0006 & 0.0004 & 0 & 0 & 0 \\
%   \hline
%\end{tabular}\label{pwkarate}
%\end{table*}

%\iffalse
\vspace{-2.8em}
\subsection{Countries data}
In this section we will test  on a direct relational data set, referred as the
benchmark data set Countries Data \cite{kaufman2009finding,mei2010fuzzy}. The task is to group twelve
countries into clusters based on the pairwise relationships as given
in Table \ref{countriestable}, which is in fact the average dissimilarity scores on some dimensions of quality of life provided
subjectively by students in a political science class. Generally,
these countries are classified into three categories: Western,
Developing and Communist. % In this experiment, the original dissimilarities between countries $i$ and $j$ are scaled by $$\tau_{ij}=d_{ij}/\max \{d_{ij}\},~~ i, j= 1,2,\cdots,12. $$
We test the performances
of FCMdd and ECMdd with two different sets of initial
representative countries which are $\Delta_1$= $\{\text{C10: USSR; C8: Israel; C7: India}\}$ and \linebreak $\Delta_2$ = $\{\text{C6: France; C4: Cuba; C1: Belgium}\}$. The three countries in $\Delta_1$ are well separated. On the contrary,  for the countries in $\Delta_2$, Belgium
is similar to France, which makes two initial medoids of three are very close in terms of the given dissimilarities. The  parameters are set as $\beta=2$ for
FCMdd, and $\beta=2, \alpha = 0.95, \eta = 1, \gamma = 1$ for ECMdd.
% latex table generated in R 3.1.1 by xtable 1.7-3 package
% Mon Feb 16 17:03:40 2015
\begin{table*}[ht]
\centering \caption{Countries data: dissimilarity matrix.}
\begin{tabular}{llllllllllllllll}
  \hline
  & Countries & C1 & C2 & C3 & C4 & C5 & C6 & C7 & C8 & C9 & C10 & C11 & C12  \\
  \hline
1 & C1: Belgium:  & 0.00 & 5.58 & 7.00 & 7.08 & 4.83 & 2.17 & 6.42 & 3.42 & 2.50 & 6.08 & 5.25 & 4.75 \\
  2 & C2: Brazil & 5.58 & 0.00 & 6.50 & 7.00 & 5.08 & 5.75 & 5.00 & 5.50 & 4.92 & 6.67 & 6.83 & 3.00 \\
  3 & C3: China & 7.00 & 6.50 & 0.00 & 3.83 & 8.17 & 6.67 & 5.58 & 6.42 & 6.25 & 4.25 & 4.50 & 6.08 \\
  4 & C4: Cuba & 7.08 & 7.00 & 3.83 & 0.00 & 5.83 & 6.92 & 6.00 & 6.42 & 7.33 & 2.67 & 3.75 & 6.67 \\
  5 & C5: Egypt & 4.83 & 5.08 & 8.17 & 5.83 & 0.00 & 4.92 & 4.67 & 5.00 & 4.50 & 6.00 & 5.75 & 5.00 \\
  6 & C6: France & 2.17 & 5.75 & 6.67 & 6.92 & 4.92 & 0.00 & 6.42 & 3.92 & 2.25 & 6.17 & 5.42 & 5.58 \\
  7 & C7: India & 6.42 & 5.00 & 5.58 & 6.00 & 4.67 & 6.42 & 0.00 & 6.17 & 6.33 & 6.17 & 6.08 & 4.83 \\
  8 & C8: Israel & 3.42 & 5.50 & 6.42 & 6.42 & 5.00 & 3.92 & 6.17 & 0.00 & 2.75 & 6.92 & 5.83 & 6.17 \\
  9 & C9: USA & 2.50 & 4.92 & 6.25 & 7.33 & 4.50 & 2.25 & 6.33 & 2.75 & 0.00 & 6.17 & 6.67 & 5.67 \\
  10 & C10: USSR & 6.08 & 6.67 & 4.25 & 2.67 & 6.00 & 6.17 & 6.17 & 6.92 & 6.17 & 0.00 & 3.67 & 6.50 \\
  11 & C11: Yugoslavia & 5.25 & 6.83 & 4.50 & 3.75 & 5.75 & 5.42 & 6.08 & 5.83 & 6.67 & 3.67 & 0.00 & 6.92 \\
  12 & C12: Zaire & 4.75 & 3.00 & 6.08 & 6.67 & 5.00 & 5.58 & 4.83 & 6.17 & 5.67 & 6.50 & 6.92 & 0.00 \\
   \hline
\end{tabular}\label{countriestable}
\end{table*}
%\fi

The results of FCMdd and ECMdd are given in
Table \ref{couFCMdd} and Table \ref{couECMdd} respectively.  It can be seen that FCMdd is very sensitive to
initializations.  When the initial prototypes are well set (the case of $\Delta_1$), the obtained partition is reasonable. However, the clustering results become worse when the initial medoids are not ideal (the case of $\Delta_2$). In fact two of the three medoids are not changed during the update process of FCMdd when using initial prototype set $\Delta_2$. This example illustrates that FCMdd is quite easy to be stuck in a
local minimum. For ECMdd, the credal partitions are the same with
different initializations. The pignistic probabilities are also displayed in Table \ref{couECMdd}, which could be regarded as membership values in fuzzy partitions. The country Egypt is clustered into imprecise class \{1, 2\}, which indicating that Egypt is not so well belongs to
Developing or Western alone, but belongs to both categories. This result is consistent with the fact shown from the dissimilarity matrix: Egypt is similar to both USA and India, but has the largest dissimilarity to China. From this experiment we could conclude that ECMdd is more robust to the initializations than FCMdd.

From Table \ref{couECMdd} we can also see the medoid of each class. For instance, China is the medoid of its cluster (Communist countries) no matter which initial prototype set is used.  This reflects the important role of China in communist countries and it has significant communist characters.
% latex table generated in R 3.1.1 by xtable 1.7-3 package
% Sun Mar 08 21:51:04 2015
\begin{table*}[ht]
\centering \caption{Clustering results of FCMdd for countries data. The prototype (medoid) of each class is marked with *.}
\begin{tabular}{llllllllllllll}
\hline
& & \multicolumn{5}{l}{FCMdd with $\Delta_1$} & &\multicolumn{5}{l}{FCMdd with $\Delta_2$}\\
\cline{3-7} \cline{9-13}
 & Countries  & $u_{i1}$ & $u_{i2}$ & $u_{i3}$ & Label & Medoids &  & $u_{i1}$ & $u_{i2}$ & $u_{i3}$ & Label & Medoids \\
  \hline
1 &        C1:    Belgium &    0.4773  &    0.2543  &    0.2685  &          1 &          -& &    1.0000  &    0.0000  &    0.0000  &          1 &          * \\

         2 &        C6:     France &    0.4453  &    0.2719  &    0.2829  &          1 &          -& &    0.0000  &    1.0000  &    0.0000  &          2 &          * \\

         3 &        C8:     Israel &    1.0000  &    0.0000  &    0.0000  &          1 &          * &  &  0.4158  &    0.3627  &    0.2215  &          1 &          - \\

         4 &        C9:        USA &    0.5319  &    0.2311  &    0.2371  &          1 &          -& &    0.4078  &    0.4531  &    0.1391  &          2 &          - \\ \\

         5 &        C3:      China &    0.2731  &    0.3143  &    0.4126  &          3 &          -& &    0.2579  &    0.2707  &    0.4714  &          3 &          - \\

         6 &        C4:       Cuba &    0.2235  &    0.2391  &    0.5374  &          3 &          -& &    0.0000  &    0.0000  &    1.0000  &          3 &          * \\

         7 &       C10:       USSR &    0.0000  &    0.0000  &    1.0000  &          3 &          * &&    0.2346  &    0.2312  &    0.5342  &          3 &          - \\

         8 &       C11: Yugoslavia &    0.2819  &    0.2703  &    0.4478  &          3 &          -& &    0.2969  &    0.2875  &    0.4156  &          3 &          - \\ \\

         9 &        C2:     Brazil &    0.3419  &    0.3761  &    0.2820  &          2 &          -& &    0.3613  &    0.3506  &    0.2880  &          1 &          - \\

        10 &        C5:      Egypt &    0.3444  &    0.3687  &    0.2870  &          2 &          -& &    0.3558  &    0.3493  &    0.2948  &          1 &          - \\

        11 &        C7:      India &    0.0000  &    1.0000  &    0.0000  &          2 &          *& &    0.3257  &    0.3257  &    0.3485  &          3 &          - \\

        12 &       C12:      Zaire &    0.3099  &    0.3959  &    0.2942  &          2 &          -& &    0.3901  &    0.3321  &    0.2778  &          1 &          - \\

   \hline
\end{tabular}\label{couFCMdd}
\end{table*}
\begin{table*}[ht]
\centering \caption{Clustering results of ECMdd for countries data. The prototype (medoid) of each class is marked with *. The Label \{1, 2\} represents the imprecise class expressing the uncertainty on class 1 and class 2.}
\begin{tabular}{llllllllllllllll}
\hline
& & \multicolumn{5}{l}{ECMdd with $\Delta_1$} & &\multicolumn{5}{l}{ECMdd with $\Delta_2$}\\
\cline{3-7} \cline{9-13}
 & Countries  & $BetP_{i1}$ & $BetP_{i2}$ & $BetP_{i3}$ & Label & Medoids &  & $BetP_{i1}$ & $BetP_{i2}$ & $BetP_{i3}$ & Label & Medoids \\
  \hline
1 & C1: Belgium & 1.0000 & 0.0000 & 0.0000 &  1 & * & & 1.0000 & 0.0000 & 0.0000 & 1 & * \\
  2 & C6: France & 0.4932 & 0.2633 & 0.2435 & 1 & - && 0.5149 & 0.2555 & 0.2297 & 1 & - \\
  3 & C8: Israel & 0.4144 & 0.3119 & 0.2738 & 1 & - && 0.4231 & 0.3051 & 0.2719 & 1 & - \\
  4 & C9: USA & 0.4503 & 0.2994 & 0.2503    & 1 & - && 0.4684 & 0.2920 & 0.2396 & 1 & - \\ \\
  5 & C3: China& 0.2323 & 0.2294 & 0.5383 & 3   & * && 0.0000 & 0.0000 & 1.0000 & 3 & * \\
  6 & C4: Cuba & 0.2778 & 0.2636 & 0.4586 & 3   & - && 0.2899 & 0.2794 & 0.4307 & 3 & - \\
  7 & C10: USSR& 0.2509 & 0.2260 & 0.5231 & 3  & -& & 0.3167 & 0.2849 & 0.3984 & 3 & - \\
  8 & C11: Yugoslavia & 0.3478 & 0.2488 & 0.4034 & 3 & -& & 0.3579 & 0.2526 & 0.3895 & 3 & - \\
 \\  9 & C2: Brazil & 0.0000 & 1.0000 & 0.0000 & 2 & *& & 0.0000 & 1.0000 & 0.0000 & 2 & * \\
  10 & C5: Egypt & 0.3755 & 0.3686 & 0.2558 &\{1, 2\} & -& & 0.3845 & 0.3777 & 0.2378 & \{1, 2\} & - \\
  11 & C7: India& 0.3125 & 0.3650 & 0.3226 & 2 & - && 0.2787 & 0.3740 & 0.3473 & 2 & - \\
  12 &  C12: Zaire & 0.3081 & 0.4336 & 0.2583 & 2 & -& & 0.3068 & 0.4312 & 0.2619 & 2 & - \\
   \hline
\end{tabular}\label{couECMdd}
\end{table*}
\subsection{UCI data sets}
\begin{center} \begin{figure}[!thbt] \centering
		\includegraphics[width=0.45\linewidth]{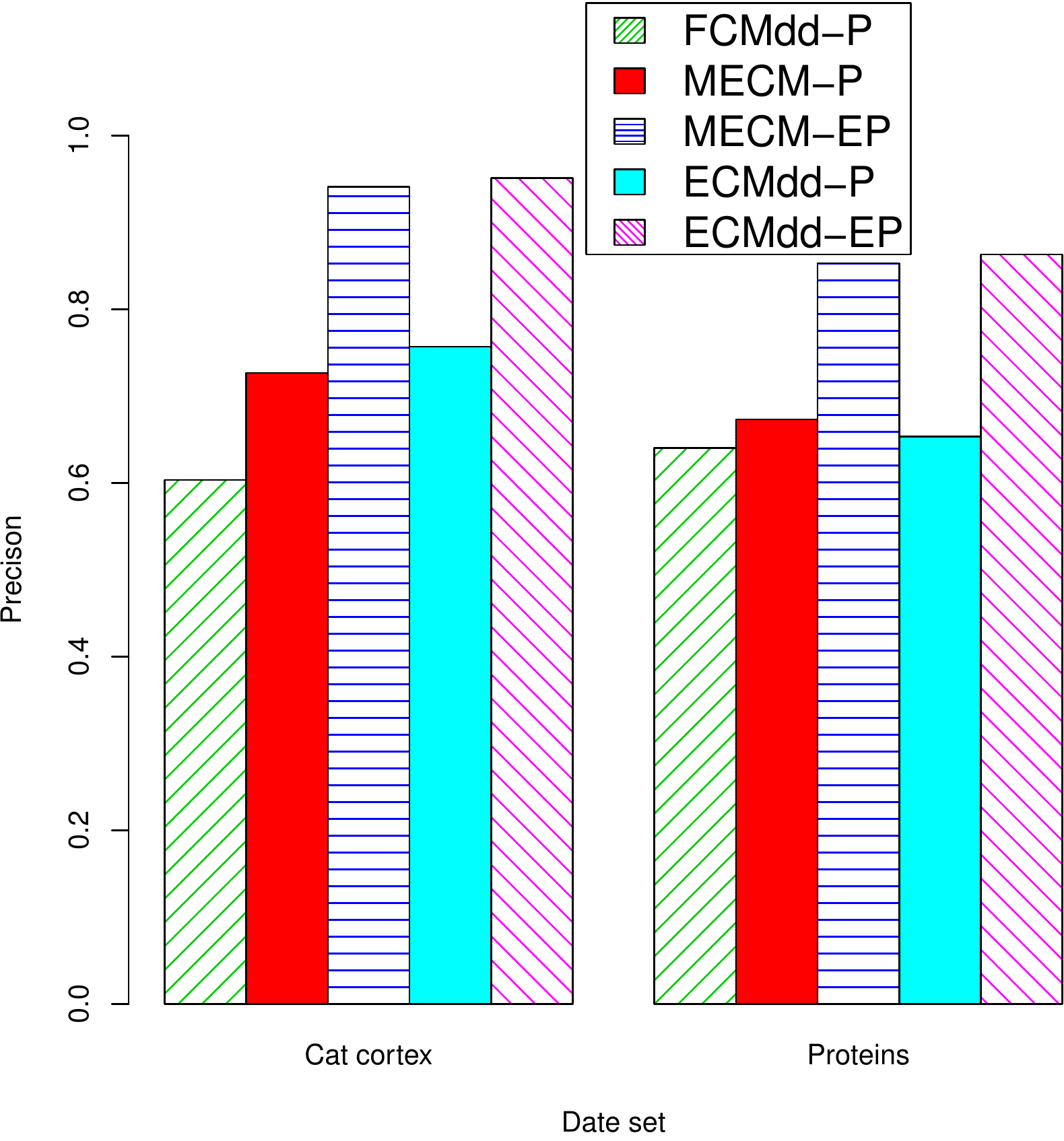}\hfill
        \includegraphics[width=0.45\linewidth]{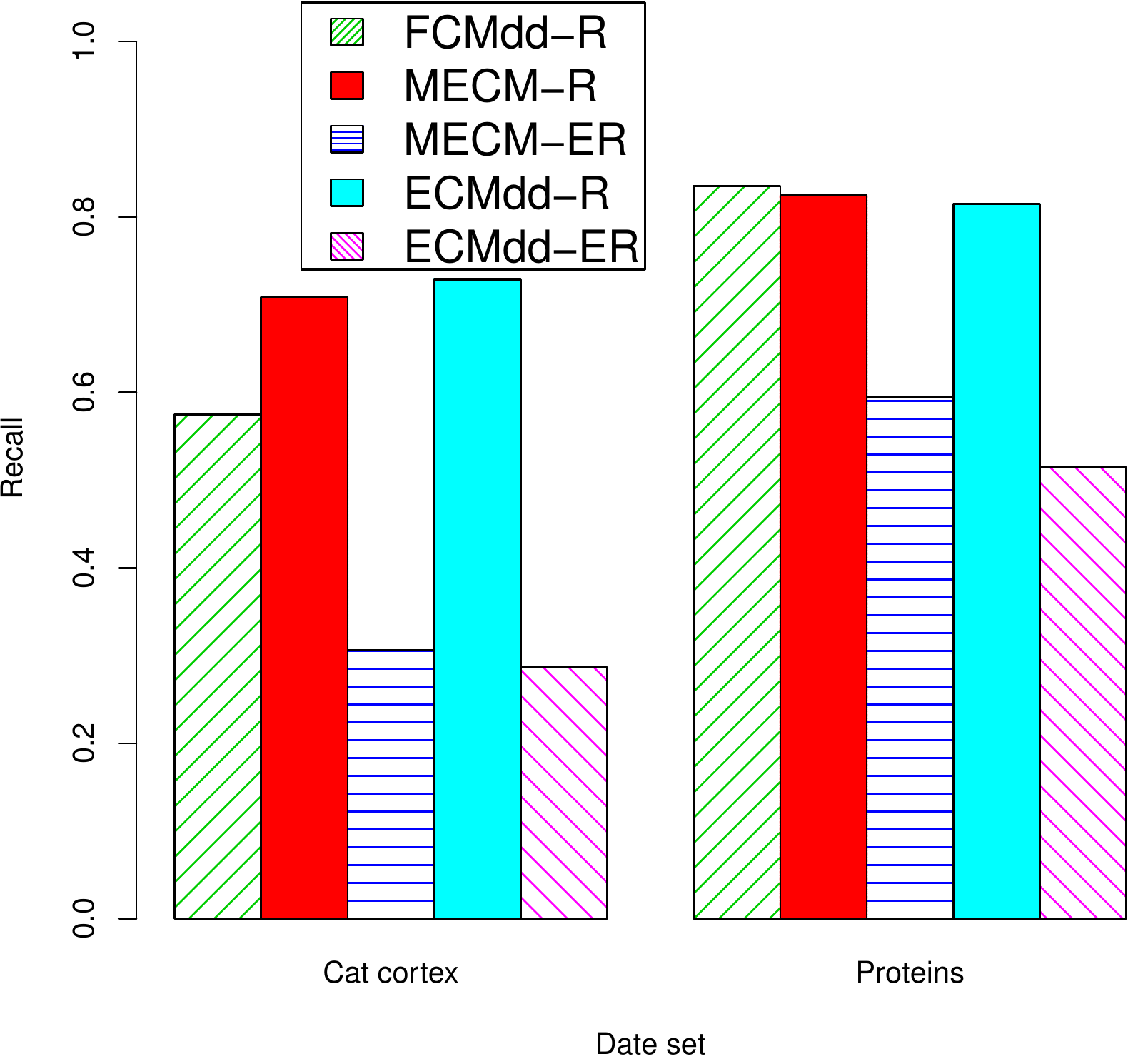} \hfill
        \parbox{.45\linewidth}{\centering\small a. Precision} \hfill
		\parbox{.45\linewidth}{\centering\small b. Recall}
		\includegraphics[width=0.45\linewidth]{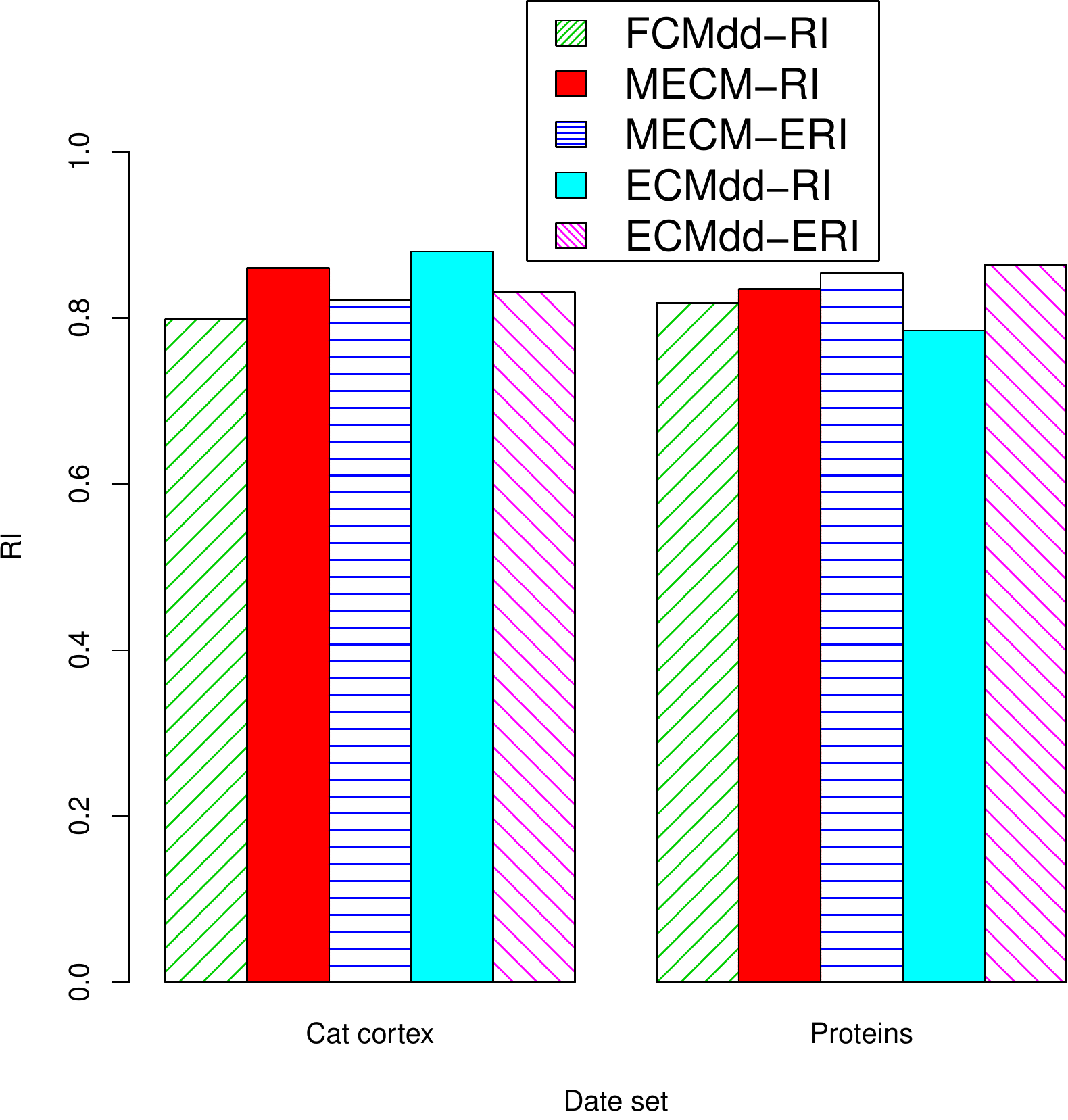}\hfill
        \parbox{.6\linewidth}{\centering\small c. RI}
\caption{The clustering results for two UCI data sets.} \label{uci} \end{figure} \end{center}

Finally the clustering performance of different methods will be compared on two benchmark UCI relational data sets: ``Cat cortex" data set and ``Protein" data set. The given information for these data sets is pair-wise relationship values. For the former it is a matrix of connection strengths between 65 cortical areas of the cat brain, while for the latter is  a dissimilarity matrix measuring
the structural proximity of 213 proteins sequences.  The comparison
results by different evaluation indices are displayed in Figure \ref{uci}. For ECMdd and MECM, the classical Precision (P), Recall (R) and Rand Index (RI) are calculated based on the pignistic probabilities, and the corresponding evidential indices are obtained from the hard credal partition \cite{masson2008ecm}. As it can be seen, the three classical measures are almost the same for all the methods. This
reflects that  pignistic probabilities  play a similar role as fuzzy
membership. But we can see that for ECMdd and MECM,  EP is significantly high. Such effect can be attributed to the introduced imprecise clusters which enable us to make a compromise decision between hard ones. But as many
points are clustered into imprecise classes, the evidential recall value is
low. The performance of ECMdd is slightly better than MECM.  But we know the expression of imprecise classes of ECMdd is more simple than that of MECM and from the experiment it proves that ECMdd is more efficient than MECM in terms of executing time.
\vspace{-2.4em}
\section{Conclusion}
In this paper, the evidential $c$-medoids  clustering is proposed as a new medoid-based
clustering algorithm.  The proposed approach is the extensions of  crisp $c$-medoids and
fuzzy $c$-medoids on the framework of belief function theory.  By the introduced imprecise clusters, we could find some overlapped and indistinguishable clusters for uncertain patterns. This results in higher accuracy of the specific decisions. The experimental results illustrates the advantages of credal partitions by ECMdd. In real applications, using only one medoid may not adequately model different types of group structure  and hence limits the clustering performance on complex data sets. Therefore, we intend to include the feature of multiple prototype representation of classes in our future research work.
\section*{Acknowledgements}
This work was supported by the National
Natural Science Foundation of China (Nos.61135001, 61403310).
%\bibliographystyle{IEEEtran}
%\bibliography{../ECMdd_paper_journal/paperlist}

\end{document}